\documentclass[runningheads]{llncs}
\usepackage{graphicx}

\usepackage{tikz}
\usepackage{comment}
\usepackage{amsmath,amssymb} 
\usepackage{color}
\usepackage{multirow}
\usepackage{cleveref}

\usepackage[accsupp]{axessibility}  

\begin{document}
\pagestyle{headings}
\mainmatter
\def\ECCVSubNumber{4705}  

\title{The Surprisingly Straightforward Scene Text Removal Method With Gated Attention and Region of Interest Generation: A Comprehensive Prominent Model Analysis} 

\titlerunning{Scene Text Removal With Gated Attention and RoI Generation}
%
\author{Hyeonsu Lee\inst{1}\orcidID{0000-0002-6317-9883}\and
Chankyu Choi\inst{1}\orcidID{0000-0002-9166-2100}
}
\authorrunning{Hyeonsu Lee et al.}
%
\institute{NAVER Corp
\email{\{hyeon-su.lee, chankyu.choi\}@navercorp.com}}
\maketitle

\begin{abstract}
Scene text removal (STR), a task of erasing text from natural scene images, has recently attracted attention as an important component of editing text or concealing private information such as ID, telephone, and license plate numbers. While there are a variety of different methods for STR actively being researched, it is difficult to evaluate superiority because previously proposed methods do not use the same standardized training/evaluation dataset. We use the same standardized training/testing dataset to evaluate the performance of several previous methods after standardized re-implementation. We also introduce a simple yet extremely effective Gated Attention (GA) and Region-of-Interest Generation (RoIG) methodology in this paper. GA uses attention to focus on the text stroke as well as the textures and colors of the surrounding regions to remove text from the input image much more precisely. RoIG is applied to focus on only the region with text instead of the entire image to train the model more efficiently. Experimental results on the benchmark dataset show that our method significantly outperforms existing state-of-the-art methods in almost all metrics with remarkably higher-quality results. Furthermore, because our model does not generate a text stroke mask explicitly, there is no need for additional refinement steps or sub-models, making our model extremely fast with fewer parameters. The dataset and code are available at \url{\textcolor{magenta}{https://github.com/naver/garnet}}.
\end{abstract}

\begin{figure}[t]
  \centering
   \includegraphics[width=0.9\linewidth]{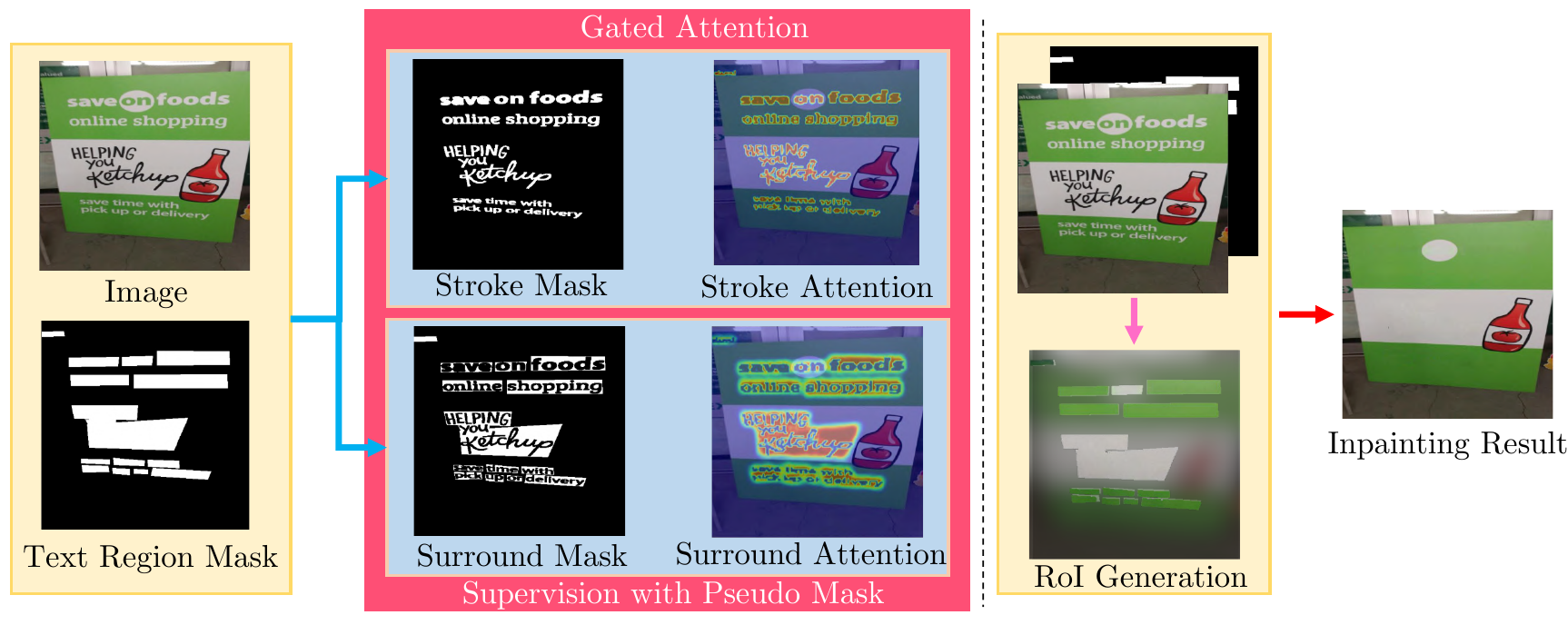}
   \caption{Visualization of the proposed model’s input and output. Visualization of the results of applying attention to the text stroke and applying attention to the text stroke surrounding regions found by the GA. We obtained satisfying STR results using RoIG.}
   \label{fig1}
\end{figure}

\section{Introduction}

\textit{Scene text removal} (STR) is a task of erasing text from natural scene images, which is useful for privacy protection, text editing in images/videos, and Augmented Reality (AR) translation.

A lot of current STR research utilizes the deep learning method. Early studies \cite{zhang2019ensnet} attempted to erase all text in an image without using a text region mask, but this produced imprecise, dissatisfactory results such as some regions being blurred out. Furthermore, it was impossible to selectively erase letters in a certain region with this method, which drastically limited applications. In order to overcome these limitations, recent studies \cite{tursun2020mtrnet++,zdenek2020erasing} in the field built text removal models after generating text region masks through manual or automatic means. This lead to comparably better quality results and wider applications.

However, these past studies did not use the same standardized training dataset and evaluation dataset, making it impossible to evaluate performance in a fair manner. For instance, some papers trained on different subsets of synthetic data\cite{gupta2016synthetic}. Furthermore, they do not have the same input image dimensions, which affects speed, accuracy, and model parameters. Thus, it is difficult to select a previously proposed model for the unique requirements of a specific application when there is no standardized comparison of the model size and speed. 

In this paper, we perform a fair comparison with both qualitative (text removal quality) and quantitative (model size, inference time) metrics by re-implementing prominent previously proposed models with the same standards, training them with the same training dataset, and evaluating them on the same evaluation dataset. Furthermore, we introduce the \textit{Gated Attention} (GA) and the \textit{Region of Interest Generation} (RoIG). GA is a simple yet extremely effective method that uses attention to focus on the text stroke and its surroundings, while RoIG drastically improves the training efficiency of the STR model.  

\Cref{fig1} shows GA found the Text Stroke Region (TSR) and the Text Stroke Surrounding Region (TSSR) after training the pseudo mask and how RoIG generates the results only for text mask regions. More details are described in \Cref{methodology}.

Our contributions can be summarized as follows:

\begin{enumerate}
\item We re-implemented previously proposed methods with the same input image size, trained them on the same training dataset, and compared their performance with the same evaluation dataset. We also performed comparisons of speed and model size.
\item We proposed a method called GA which not only distinguishes between the background and the text stroke, but also utilizes attention to identify both TSR and TSSR. To the best of our knowledge, there is no other previous study that considered both TSR and TSSR while performing STR. 
\item  We proposed a method called RoIG, which helps the network focus on only the region with text instead of the entire image to train the model more efficiently.
\item The proposed method generates higher quality results than existing state-of-the-art methods on both synthetic and real datasets. It is also significantly lighter and faster than most other prominent previously proposed popular methods because our model does not have additional refinement steps or sub-models needed to generate a text stroke mask.
\end{enumerate}

\section{Related Works}

\setlength{\tabcolsep}{4pt}
\begin{table*}[t]
\centering
\caption{A side-by-side comparison of how our proposed model differs from previous works.}
{
\begin{tabular}{lcccccc}
\hline
\multirow{2}{*}{Method} & Use & Selective & Stage & Stroke & Surround  & RoI \\
 & text box &removal &  & localization & localization & Generation\\
\hline
EnsNet \cite{zhang2019ensnet} & x & x & 1 & x & x & x \\
MTRNet \cite{tursun2019mtrnet} & o & o & 1 & x & x & x\\
MTRNet++ \cite{tursun2020mtrnet++} & o & o & 2 & o & x & x \\
EraseNet \cite{liu2020erasenet} & x & x & 2 & x & x & x \\
Tang \textit{et al.} \cite{tang2021stroke} & o & o & 2 & o & x & o(crop) \\
\hline
Ours & o & o & 1 & o & o & o\\
\hline
\end{tabular}
}
\label{table1}
\end{table*}
\setlength{\tabcolsep}{1.4pt}

The major trend in scene text removal before the emergence of deep learning was traditional rule-based methods  \cite{bertalmio2001navier,telea2004image}, which are often hand-crafted and require prior domain knowledge.

Recently, deep learning-based text removal has been proposed by adopting popular GAN-based methods. EnsNet \cite{zhang2019ensnet} is simple and fast because it doesn't need any auxiliary inputs. However, its results are blurry and of low quality. Furthermore, its practical applications are limited because it is impossible to only erase text in a certain region.
MTRNet \cite{tursun2019mtrnet} requires the generation of text box region masks through manual or automatic means. These studies show that text region masks can improve the network’s performance but cannot guarantee high-quality results.
MTRNet++ \cite{tursun2020mtrnet++} and EraseNet \cite{liu2020erasenet} proposed a coarse-refinement two-stage network. While the results are of higher quality, the model is much bigger, slower, and too complicated.

Zdenek \textit{et al.} \cite{zdenek2020erasing} used an auxiliary text detector to retrieve the text box mask, then attempted to erase text through a general inpainting method. However, they were unable to generate results of satisfactory quality because they did not consider qualities specific to text like the text strokes. Tang \textit{et al.} \cite{tang2021stroke} can train their model in a rather efficient manner because they erase text by cropping only the text regions. However, this method ignores all global contexts other than the cropped region and has difficulty precisely cropping the region of curved texts.

We designed a model that can focus on only the text region area while also managing to take the global context of the entire image into consideration. This model does not require an additional refinement process or a sub-model dedicated to text stroke localization, leading to a drastically faster and lighter model.

\setlength{\tabcolsep}{4pt}
\begin{table*}[t]
\centering
\caption{A comparison of our model’s performance with previous STR models on real data (SCUT-EnsText \cite{liu2020erasenet}). We compare the results claimed in the original papers of the previous prominent STR models with the results we re-measured in an equal environment. We realize that there are fundamental differences in models and that it would be unfair to evaluate them under potentially disadvantageous circumstances. To make things truly fair, we also pasted the text region (produced by the model using a text box mask) over the original image, then evaluated the results yet again. The values right of “/” reflect experiments conducted after this adjustment was made. Our model produces superior results in all metrics regardless. The best score is highlighted in bold.}
{
\begin{tabular}{@{}p{0.1em}@{\quad} l|l|c|ccc}
\hline
\multirow{2}{*}{} & \multirow{2}{*}{Method} & Train & \multirow{2}{*}{Size} & \multicolumn{3}{c}{Image-Eval}  \\
      &   &  Data  &  & PSNR   & SSIM   & AGE \\
\hline
\multirow{5}{*}{\rotatebox[origin=c]{90}{Reported}} & Scene Text Eraser \cite{nakamura2017scene} & Real & 256 & 25.47/-    &   90.14/-      &  6.01/- \\
& EnsNet \cite{zhang2019ensnet} & Real &   512   &  29.53/-    &   92.74/-     & 4.16/- \\
& MTRNet \cite{tursun2019mtrnet} &Syn(75\%) & 256 & -/- & -/-  & -/- \\
& MTRNet++ \cite{tursun2020mtrnet++} & Syn(95\%) & 256 & -/- & -/- & -/- \\
& EraseNet \cite{liu2020erasenet} & Real & 512 & \textbf{32.30/37.26} & \textbf{95.42/96.86} & \textbf{3.02/-} \\
\hline
\multirow{7}{*}{\rotatebox[origin=c]{90}{Our experiment}} & EnsNet \cite{zhang2019ensnet} & Real+Syn & 512 & 31.05/32.99 & 94.78/95.16 & 2.67/1.85 \\
& MTRNet \cite{tursun2019mtrnet} & Real+Syn & 256 & 30.61/36.06 & 89.85/95.72 & 3.92/1.21 \\
 & & Real+Syn & 512 & 32.46/36.89 & 95.86/96.41 & 3.12/0.97  \\
& MTRNet++ \cite{tursun2020mtrnet++} & Real+Syn & 256 & 35.29/37.40 & 96.31/96.68  & 1.26/1.09 \\
 & & Real+Syn & 512 & 34.86/36.50 & 96.32/96.51  & 1.48/1.35 \\
& EraseNet \cite{liu2020erasenet} & Real+Syn & 512 & 30.54/37.16 & 96.27/97.53  & 3.07/1.20 \\
& EraseNet \cite{liu2020erasenet} + M & Real+Syn & 512 & 34.29/40.18 & 97.73/97.98  & 2.28/0.69 \\
\cline{2-7}
& Ours & Real+Syn & 512 & \textbf{-/41.37} & \textbf{-/98.46}  & \textbf{-/0.64} \\
\hline
\end{tabular}
}
\label{table2}
\end{table*}
\setlength{\tabcolsep}{1.4pt}

\section{Comprehensive Prominent Model Analysis}\label{modelanalysis}
In this section, we analyze the difference between previous methods and ours. We also show the results of evaluating previous methods with one standardized dataset.

\Cref{table1} outlines the specific differences between our method and previously proposed methods. First, our proposed method takes a text box mask as the input to our model, leading to results of significantly better quality as well as the option for users to selectively erase only the text that they wish to. Second, our proposed method can localize the TSR and TSSR to erase text in a surgical manner. Finally, the use of RoIG makes our proposed method show off significantly better results than previous methods without even implementing a coarse-refinement 2-stage network. 

\Cref{table2}, \Cref{table3} and \Cref{table4} show the performance of several previous STR methods on real and synthetic data. Details of the experiment and re-implementation are in \Cref{experiment} and \Cref{Re-implementation details}. 

After performing an objectively fair comparison, we found that our method generates higher quality results than existing state-of-the-art methods on both synthetic and real datasets. It is also significantly lighter and faster than any other method except EnsNet \cite{zhang2019ensnet}.

\setlength{\tabcolsep}{4pt}
\begin{table*}[t]
\centering
\caption{A comparison of our model’s performance with previous STR models on real data (SCUT-EnsText \cite{liu2020erasenet}). R, P, and F refer to recall, precision and F-score, respectively. The best score is highlighted in bold.}
{
\begin{tabular}{@{}p{0.1em}@{\quad} l|l|c|ccc|rr}
\hline
\multirow{2}{*}{} & \multirow{2}{*}{Method} & Train & \multirow{2}{*}{Size} & \multicolumn{3}{c}{Detection-Eval} & GPU & \multirow{2}{*}{Params}\\
      &   &  Data  &  & P   & R   & F & Time(ms) & \\
\hline
\multirow{5}{*}{\rotatebox[origin=c]{90}{Reported}} & Scene Text Eraser \cite{nakamura2017scene} & Real & 256 & 40.9 & 5.9 &  10.2 & & \\
& EnsNet \cite{zhang2019ensnet} & Real &  512  &  68.7  &  32.8  & 44.4 & - & 12.4M\\
& MTRNet \cite{tursun2019mtrnet} &Syn(75\%) & 256 & - & -  & - & - & 54.4M\\
& MTRNet++ \cite{tursun2020mtrnet++} & Syn(95\%) & 256 & - & - & - & - & 18.7M \\
& EraseNet \cite{liu2020erasenet} & Real & 512 & 53.2 & 4.6 & 8.5 & - & - \\
\hline
\multirow{7}{*}{\rotatebox[origin=c]{90}{Our experiment}} & EnsNet \cite{zhang2019ensnet} & Real+Syn & 512 & 73.1 & 54.7 & 62.6 & 12.0 & 12.4M \\
& MTRNet \cite{tursun2019mtrnet} & Real+Syn & 256 &  &  & & 21.9 & 50.3M \\
 & & Real+Syn & 512 & 69.8 & 41.1 & 51.2 & 51.3 &  \\
& MTRNet++ \cite{tursun2020mtrnet++} & Real+Syn & 256 &  &   & & 69.8 & 18.7M \\
 & & Real+Syn & 512 & 58.6 & 20.5  & 30.4 & 238.7 & \\
& EraseNet \cite{liu2020erasenet} & Real+Syn & 512 & 40.8 & 6.3  & 10.9 & 47.4 & 17.8M \\
& EraseNet \cite{liu2020erasenet} + M & Real+Syn & 512 & 37.3 & 6.1  & 10.3 & 47.4 & 17.8M \\
\cline{2-9}
& Ours & Real+Syn & 512 & \textbf{15.5} & \textbf{1.0}  & \textbf{1.8} & \textbf{14.9} & \textbf{12.4M}\\
\hline
\end{tabular}
}
\label{table3}
\end{table*}
\setlength{\tabcolsep}{1.4pt}
\setlength{\tabcolsep}{4pt}
\begin{table*}
\centering
\caption{A comparison of our model’s performance with previous STR models on synthetic data (Oxford \cite{gupta2016synthetic}). The notation is the same as \Cref{table2} and \Cref{table3}.}
{
\begin{tabular}{l|c|ccc|ccc}
\hline
\multirow{2}{*}{Method} & Input & \multicolumn{3}{c}{Image-Eval} & \multicolumn{3}{c}{Detection-Eval} \\
                      & Size & PSNR   & SSIM    & AGE  & P & R &  F \\
                        \hline
Original Image & 512 & & &  & 71.3 & 51.5 & 59.8 \\
\hline
EnsNet \cite{zhang2019ensnet} &512 & 36.67/39.74& 97.71/97.94  & 1.25/0.77 & 55.1 & 14.0 & 22.3 \\
MTRNet \cite{tursun2019mtrnet} & 256 & 30.96/37.69 & 90.95/95.83  & 4.17/1.22 \\
		& 512 &  35.49/40.03 & 97.10/97.69  & 2.18/0.80 & 58.6 & 13.7 & 22.2 \\
MTRNet++ \cite{tursun2020mtrnet++} & 256 & 37.40/40.26 & 97.02/97.25  & 0.86/0.79  \\
		& 512 &  38.31/40.64 & 97.82/97.94  & 0.81/0.73 & 64.0 & 15.9 & 25.4 \\
EraseNet \cite{liu2020erasenet} &   512 & 34.35/41.73 & 98.01/98.62  & 1.81/0.66 & 30.8 & 0.6 & 1.2 \\
EraseNet \cite{liu2020erasenet} + M &   512 & 36.37/42.98 & 98.50/\textbf{98.75}  & 1.72/0.56 & 31.4 & \textbf{0.0} & 1.4 \\
\hline
Ours &   512 & \textbf{-/43.64} & -/98.64 & \textbf{-/0.55} & \textbf{18.9} & 0.1 & \textbf{0.3} \\
\hline
\end{tabular}
}
\label{table4}
\end{table*}
\setlength{\tabcolsep}{1.4pt}

\section{Methodology}\label{methodology}
\subsection{Motivation}
Previous STR models \cite{tursun2019mtrnet,tursun2020mtrnet++} used a text box region as well as a text stroke region in an attempt to perform precise text removal. However, TSSR was mostly overlooked. After finding inspiration from observing how humans must alternate between paying attention to the text stroke regions and the surrounding regions of the text while manually performing STR, we devised the GA. Meanwhile, because all previous studies performed STR on the entire image, artifacts frequently occurred in non-text regions. Thus, we devised the RoIG, which allows our STR model to only generate a result image from within the text box region instead of wasting resources attempting to perform STR on the full image.

\begin{figure*}[t]
\centering
\includegraphics[width=1\textwidth]{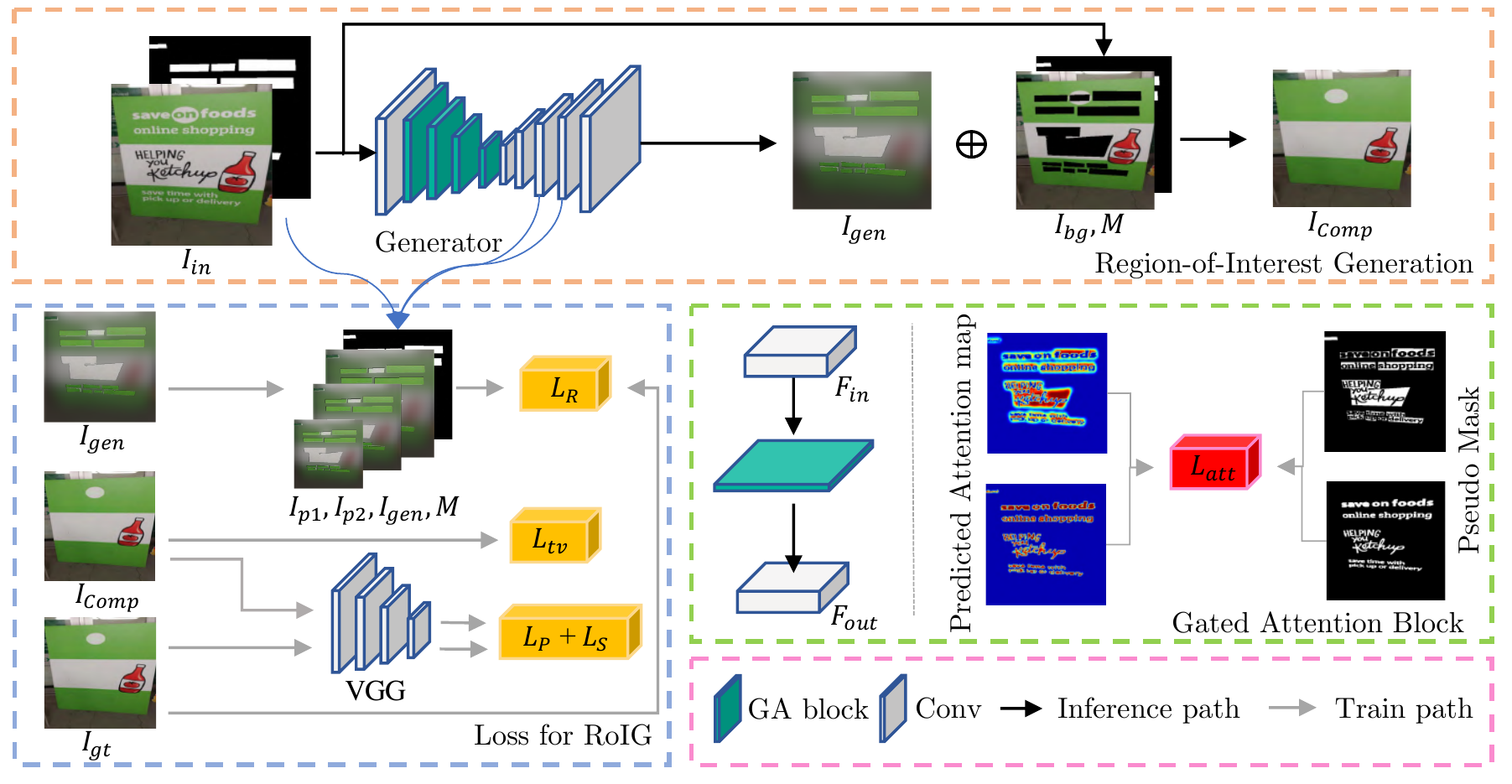}
\caption{The overall architecture of the proposed model. The top box, Region-of-Interest Generation, shows that our model generates images focusing only on the text region. The box on the bottom left, Loss for RoIG, how our loss function helps the model focus only on the text region. The $L_R$ only use $I_{gen}$ with $M$, and $L_{tv}$, $L_p$ and $L_s$ use $I_{comp}$, so the outside of text regions in $I_{gen}$ does not participate in loss calculation. On bottom right box, The GA calculates the stroke attention and stroke surrounding attention mask for every layer of the generator’s encoder, then automatically calculates the importance of each. Each attention map is supervised by pseudo mask. More detailed information on GA is in \Cref{fig3}.}
\label{fig2}
\end{figure*}

\subsection{Model architecture}
\Cref{fig2} shows the architecture of our proposed method. The generator G takes the image and corresponding text box mask as its input and produces a non-text image that is visually plausible. Following GAN-based methods \cite{mirza2014conditional,isola2017image}, the discriminator D takes both the input of the generator and the target images as its input and differentiates between real images and images produced by the generator. The objective functions of the generator and discriminator are as follows:
\begin{equation}\label{eq1}
L_{adv} = \mathbb{E}_{x} \left[ log D(x,G(x)) \right]
\end{equation}
\begin{equation}\label{eq2}
\begin{split}
L_{D} = \mathbb{E}_{x,y} \left[ log D(x,y) \right] + \mathbb{E}_{x} \left[(1 - log D(x,G(x))) \right]
\end{split}
\end{equation}
where $x$ is the input image concatenated with a text box mask and y is the target image.

\textbf{Generator}.
The generator has an FCN-ResNet18 backbone and skip connections \cite{ronneberger2015u} between the encoder and decoder. The model is composed of five convolution layers paired with five deconvolution layers with a kernel size of 4, stride of 2, and padding of 1. The convolution pathway is composed of two residual blocks \cite{he2016deep}, which contains the proposed Gated Attention (GA) module.

\textbf{Discriminator}.
For training, we use a local-aware discriminator proposed in EnsNet \cite{zhang2019ensnet}, which only penalizes the erased text patches. The discriminator is trained with locality-dependent labels, which indicate text stroke regions in the output tensor. It guides the discriminator to focus on text regions.

\subsection{Gated Attention}
Localizing both the TSR and TSSR is imperative to perform surgical STR. In order to do this without letting the size of our model blow up, we used spatial attention\cite{woo2018cbam} instead of a separate image segmentation branch\cite{tursun2020mtrnet++}. \Cref{table2} shows how our proposed model is significantly faster and smaller than MTRNet++ \cite{tursun2020mtrnet++}.

\begin{figure}[t]
  \centering
   \includegraphics[width=0.5\linewidth]{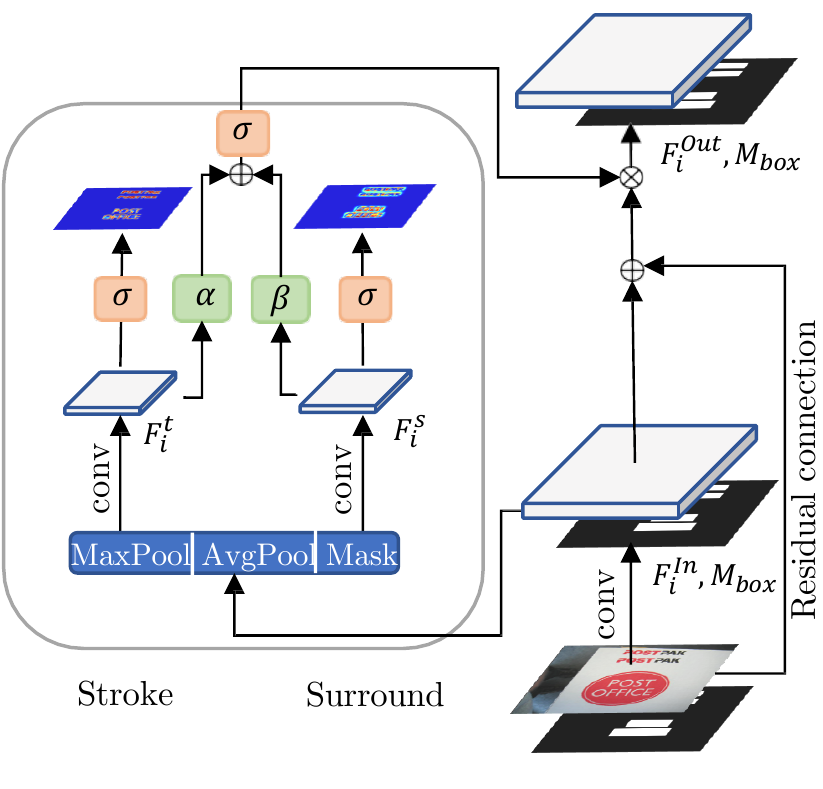}
   \caption{Architecture of the Gated Attention. The GA calculates the TSR and TSSR attention masks, then automatically calculates the importance of each through gate parameters $\alpha,\beta$.}
   \label{fig3}
\end{figure}

\Cref{fig3} shows the architecture of the GA. The module takes the feature map as its input and generates a TSR and TSSR feature map, then adjusts the proportion of these two feature maps through gate parameters.
The process of GA is as follows:
\begin{equation}\label{eq3}
F_{i}^{'} = (MaxPool(F_{i}^{In}) \oplus AvgPool(F_{i}^{In}) \oplus M_{box})
\end{equation}
\begin{equation}\label{eq4}
\begin{split}
F_{i}^{t} = W_{i}^t\cdot F_{i}^{'} \\
F_{i}^{s} = W_{i}^s\cdot F_{i}^{'}
\end{split}
\end{equation}
\begin{equation}\label{e5}
A_{i}^{out} = \sigma(\alpha_{i} F_{i}^{t} + \beta_{i} F_{i}^{s})
\end{equation}
\begin{equation}\label{eq6}
F_{i}^{out} = F_{i}^{In}A_{i}^{out}
\end{equation}
where $i$ denotes the $ith$ layer in the encoder, $F^{In}$ and $F^{out}$ denote the input and output feature maps, $W^t$ and $W^s$ denote 7x7 convolution filters that extract TSR and TSSR features, $F^{t}$ and $F^{s}$ denote extracted feature maps for localized TSR and TSSR, $\alpha$ and $\beta$ denote gate parameters, and $\sigma$ and $A^{out}$ denote the Sigmoid activation function and the attention score map.

\Cref{fig1} shows the text box mask, TSR mask, and the TSSR mask necessary for this. We generated a pseudo text stroke mask, automatically calculated by taking the pixel value difference between the input image and ground truth image. The TSR masks help train the TSR attention module to distinguish the TSR. The TSSR mask is the intersection region between the exterior of the pseudo text stroke mask and the interior of the text box mask. It makes the TSSR attention module train to focus on the colors and textures of the TSSR. Note that the attention method that we used differs from S. Woo \textit{et al.} \cite{woo2018cbam}, which was trained in a weakly-supervised manner. We show that our method is superior in \Cref{ablation}. The GA learns the gate parameter on its own, and can thus adjust the respective attention ratios allocated to TSR and TSSR during training. The loss function is designed to be applied only within the text box regions, not to the entire score map. The loss function to train GA is as follows:
\begin{equation}\label{eq7}
\begin{split}
L_{att}^{t} = \begin{cases} -Gt_i^t log(S_i^t) & \mbox{if }M_{i}^{Box} > 0, \\
0 & \mbox{otherwise}
\end{cases} \\
L_{att}^{s} = \begin{cases} -Gt_i^s log(S_i^s) & \mbox{if }M_{i}^{Box} > 0, \\
0 & \mbox{otherwise}
\end{cases}
\end{split}
\end{equation}
\begin{equation}\label{eq8}
L_{att} = L_{att}^{t} + L_{att}^{s}
\end{equation}
where $Gt_{i}^{t}$ and $Gt_{i}^{s}$ are the $ith$ pixel of the ground truth mask representing TSR and TSSR. $S_{i}^{t}$ and $S_{i}^{s}$ are the $ith$ pixel of the TSR and TSSR attention score maps. $M_{i}^{Box}$ is text box mask. The shape of $M$, $Gt$ and $S$ is $(H_n, W_n, 1)$. $H_n$ and $W_n$ denote the sizes of feature map in the $nth$ layer.

\subsection{Region-of-Interest Generation}
Most STR methods attempt to perform in-painting of TSR as well as reconstruction of the entire image. However, our approach can skip the reconstruction of non-masked regions altogether because if the text box region is given as an input to the STR model, there is no need to render the entire image for the output. Therefore, we modified the loss function so that our model’s generator only has to focus on the text box region during training. Note that the generator’s loss is only calculated with respect to the text box region. Every other region is considered \textit{don't care} and therefore is irrelevant during training. Because all regions other than the text box obtained from the generator’s output value are not used, we blurred them in Fig 3. This makes training significantly easier.

In total, we modified four loss functions for RoIG: RoI Regression, Perceptual, Style, and Total Variation Loss.

\textbf{RoI Regression Loss}. We modified regression loss to only consider text regions.
The proposed \textit{Region-of-Interest Regression Loss} is defined as:
\begin{equation}\label{eq9}
L_R(M,I_{out},I_{gt}) = \sum_{i=1}^n\lambda_i \odot M_i\parallel{I_{out(i)}-I_{gt(i)}}\parallel_1 
\end{equation}
where $I_{out(i)}$ is the output of the $ith$ deconvolution pathway. We use the output of the 3rd, 4th, and 5th layers in the deconvolution pathway. $M_i$ and $I_{gt(i)}$ are the box mask and ground truth that was resized to the same scale as $I_{out(i)}$. $\lambda_i$ is the weight for scale. We set $\lambda_i$ to 0.6, 0.8, 1.0.

\textbf{Perceptual Loss}. Perceptual Loss \cite{johnson2016perceptual} is used to make the generated image more realistic. It reduces the difference between the high-level features of the two images extracted using the pre-trained ImageNet. Some works \cite{liu2018image,zhang2019ensnet,liu2020erasenet} used a composited output to address the discrepancy between text-erased regions and the background. We use only the text regions in the generated image by pasting it into the input image. We address the discrepancy between the text box and input images by designing the loss function to use two composited outputs generated using the box and stroke mask, respectively. Perceptual Loss is defined as:
\begin{equation}\label{eq10}
\begin{aligned}
I_{boxComp} = I_{in}(1-M_{Box}) + I_{out}M_{Box} \\
I_{StrokeComp} = I_{in}(1-M_{Stroke}) + I_{out}M_{Stroke} 
\end{aligned}
\end{equation}
\begin{equation}\label{eq11}
\begin{split}
L_P = \sum_{n=1}^{N-1}\parallel{A_n(I_{boxComp}) - A_n(I_{gt})}\parallel_1 \\+ \sum_{n=1}^{N-1}\parallel{A_n(I_{StrokeComp}) - A_n(I_{gt})}\parallel_1
\end{split}
\end{equation}
where $I_{in}$, $I_{out}$ refer to the input image and the generator's output image, $I_{BoxComp}$, $I_{StrokeComp}$ are generated images composited of box mask $M_{box}$ and stroke mask $M_{stroke}$ respectively. $I_{gt}$ is the ground truth image, and $A_n$ refers to the activation of the $nth$ layer in network. We use the pool1, pool2 and pool3 layers of the VGG-16 \cite{simonyan2014very} pretrained on ImageNet \cite{deng2009imagenet}.

\textbf{Style Loss}. Style Loss \cite{gatys2015neural} considers the global texture of the entire image and is used to further improve the visual quality of the output. It is calculated using the Gram matrix of feature maps. Like perceptual loss, we use two composited outputs. Style loss is defined as: 
\begin{equation}\label{eq12}
\begin{split}
L_{S} = \sum_{n=1}^{N-1}\parallel{ \frac{1}{H_nW_nC_n}[(\phi(A_n(I_{boxComp}))-\phi(A_n(gt))]}\parallel_1 \\
+ \sum_{n=1}^{N-1}\parallel{\frac{1}{H_nW_nC_n}[(\phi(A_n(I_{strokeComp}))-\phi(A_n(gt))]}\parallel_1
\end{split}
\end{equation}
where $\phi(x) = x^Tx$ is a gram matrix operator and $A_n$ is the activation of the $nth$ layers of VGG-16 \cite{simonyan2014very}. We use same layers as Perceptual Loss. 

\textbf{Total Variation Loss}. J. Johnson \textit{et al.} \cite{johnson2016perceptual} proposed total variation loss for global denoising. As our model generates images using RoIG method, the loss function uses composited images which are generated using box masks. The total variation loss is as follows:
\begin{equation}\label{eq13}
\begin{split}
L_t = \sum_{i,j}\parallel{I^{i,j+1}_{Comp}-I^{i,j}_{Comp}}\parallel_1 \\
+ \parallel{I^{i+1,j}_{Comp}-I^{i,j}_{Comp}}\parallel_1
\end{split}
\end{equation}
where $i$,$j$ are the pixel positions.

\textbf{Total Loss function}.
The total loss function for train generator is as follows:
\begin{equation}\label{eq14}
\begin{split}
L_G = \lambda_rL_R + \lambda_pL_P + \lambda_sL_S + \lambda_tL_t \\ + \lambda_{adv}L_{adv} + \lambda_{att}L_{Att}
\end{split}
\end{equation}
where $\lambda_r$ to $\lambda_{att}$ represent the weight of each loss. We set $\lambda_i$ to 100, 0.5, 50.0, 25.0, 1, 10.

\section{Experiment}\label{experiment}
\subsection{Dataset and Evaluation Metrics}
For training and evaluation, we use both synthetic and real datasets.

\textbf{Synthetic data}. The Oxford Synthetic text dataset \cite{gupta2016synthetic} is adopted for training and evaluation. The dataset contains around 800,000 images composed of 8,000 text-free images. We randomly selected 95\% images for training, selected 10,000 images for testing, and used the rest for validation. Note that the background images in the train set and test set are mutually exclusive.

\textbf{Real data}. SCUT-EnsText \cite{liu2020erasenet} is a real dataset for scene text removal. The dataset, which was manually generated from Chinese and English text images, contains 2,749 train and 813 test images. In this paper, we adopted these images for training and evaluation.

\textbf{Preprocessing}. We need stroke-level segmentation masks to train our model. However, the existing datasets do not provide stroke-level segmentation masks. Therefore, we created it automatically by calculating the pixel value difference between the input image and the ground truth image. To suppress noise, we set a threshold of 25.

We combined synthetic and real datasets. In total, we used 738,113 images for training. The test set was used separately to distinguish between performance on real and synthetic datasets.

\textbf{Evaluation Metrics}
T. Nakamura \textit{et al.} \cite{nakamura2017scene} proposed an evaluation method using an auxiliary text detector. An auxiliary text detector obtains detection results on the images with text removed. Then, it evaluates the model performance by calculating Precision, Recall, and F-score values. A lower value means that the texts are better erased. In this paper, we use Detection Eval \cite{WolfIJDAR2006} as an evaluation metric and CRAFT \cite{baek2019character} as an auxiliary detector. However, that method only indicates how much text has been erased, not output quality. S. Zhang \textit{et al.} \cite{zhang2019ensnet} proposed using the evaluation method that is used in image inpainting. They used PSNR, SSIM, MSE, AGE, pEPs, pCEPs to evaluate image quality. The higher the value of PSNR and SSIM, and the lower the value of other metrics, the better the quality of the output image. We use PSNR, SSIM, and AGE for evaluation.

\subsection{Implementation details}
We trained our model for 6 epochs with batch size 30 on the combined dataset. The Adam optimizer with $\beta$ (0.9, 0.999) was used. We set the initial learning rate to 0.0005 and divided it by 5 every 50,000 steps. PyTorch and NVIDIA Tesla M40 GPUs were used in all experiments.

\subsection{Re-implementation details}\label{Re-implementation details}
In this section, we provide details of the re-implementation of previous methods.
We re-implemented EnsNet \cite{zhang2019ensnet} by modifying the code implemented in mxnet and trained the model with the same hyperparameters used in their paper.
We re-implemented MTRNet \cite{tursun2019mtrnet} by converting the code implemented in TensorFlow to PyTorch and trained the model with the same batch size and epoch as mentioned in their paper. 
We used the official implementation of MTRNet++ \cite{tursun2020mtrnet++} and EraseNet \cite{liu2020erasenet}. We trained them with the same batch size and epoch as mentioned in their respective papers. EraserNet \cite{liu2020erasenet} + M refers the model using mask as its input with image.

\subsection{Ablation Study}\label{ablation}
In this section, we validate the effectiveness of our contributions: Gated Attention (GA) and Region-of-Interest Generation (RoIG)

\textbf{BaseLine}. We combined the text region mask with the EnsNet \cite{zhang2019ensnet} model to make a baseline. This results in a quality improvement of the output image as well as the added functionality of flexibly removing only specific characters at the user’s discretion. All of our experiments including the baseline use a 4-channel input by concatenating the 3-channel RGB and a 1-channel mask.

\setlength{\tabcolsep}{4pt}
\begin{table*}[t]
\caption{Result quality comparison with ablation studies. SA, TSRA, TSSRA, GA, and RoIG refer to Simple Attention \cite{woo2018cbam}, Text Stroke Region Attention, Text Stroke Surrounding Region Attention, Gated Attention, and Region of Interest Generation respectively. The notation is the same as \Cref{table2}.}
\centering
{\footnotesize
\begin{tabular}{l|ccc|ccc}
\hline
\multirow{2}{*}{Method} & \multicolumn{3}{c}{Image Eval} & \multicolumn{3}{c}{Detection Eval} \\
                        & PSNR   & SSIM     & AGE  & P          & R         & F         \\
                        \hline
Original Image &  &  & & 79.8  &   67.2  & 73.0   \\
\hline
EnsNet \cite{zhang2019ensnet} &   31.05/32.99 & 94.78/95.16  & 2.67/1.85 & 73.1 & 54.7 & 62.6  \\
\hline
BaseLine (EnsNet + M) &  35.65/37.28 & 96.67/96.78  & 1.50/0.89 & 63.3 & 33.2 & 43.6 \\
\hline
\hline
BaseLine + SA & 35.73/37.24 & 96.53/96.62 & 1.52/0.89 & 65.5 & 34.4 & 45.1 \\
\hline
BaseLine + TSRA &   36.07/38.06 & 97.23/97.34 & 1.46 /0.84 & 54.3 & 21.1 & 30.4 \\
\hline
BaseLine + TSSRA &   36.12/38.45 & 97.29/97.46 & 1.51/0.84 & 39.1 & 6.0 & 10.5  \\
\hline
BaseLine + GA &   36.38/38.82 & 97.56/97.72  & 1.46/0.80 & 30.6 & 3.9 & 6.9 \\
\hline
\hline
BaseLine + RoIG &  -/40.82 & -/98.19  & -/0.66 & 27.5 & 3.1 & 5.6 \\
\hline
BaseLine + GA + RoIG & \textbf{-/41.37} & \textbf{-/98.46}   &  \textbf{-/0.64} & \textbf{15.5} & \textbf{1.0} & \textbf{1.8} \\
\hline
\end{tabular}
}
\label{table5}
\end{table*}
\setlength{\tabcolsep}{1.4pt}

\begin{figure*}[t]
\centering
\includegraphics[width=0.9\textwidth]{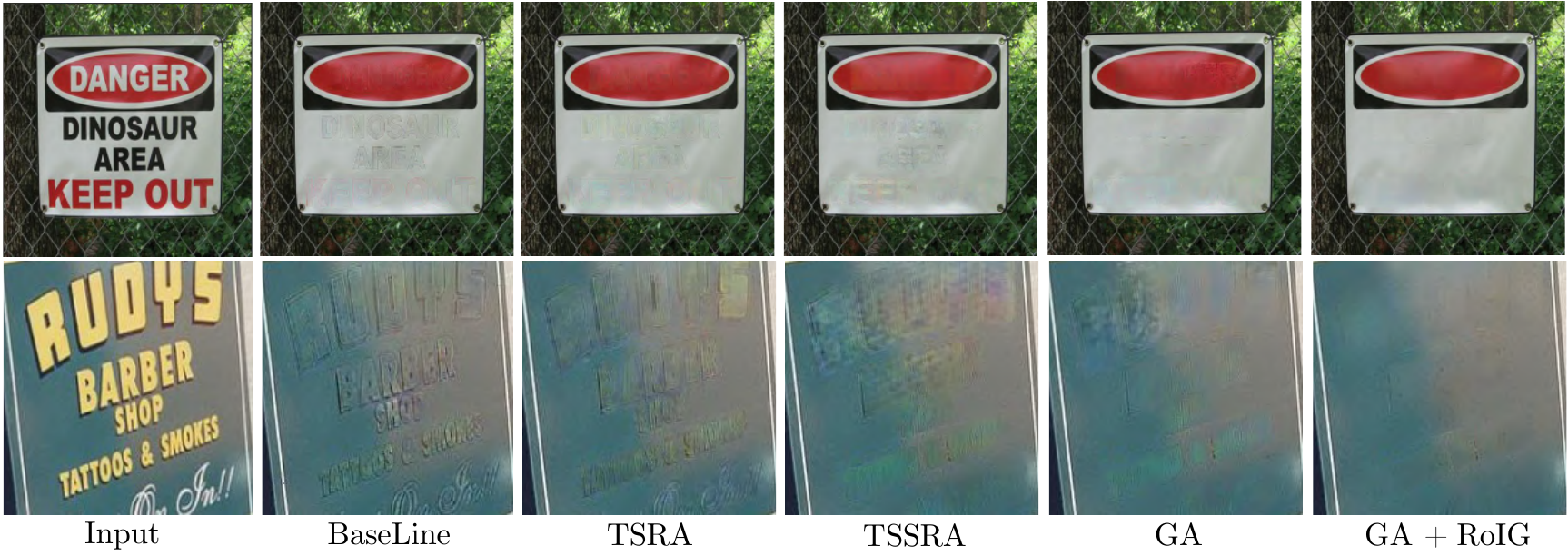} 
\caption{Comparison of the output results after ablation. Image from left to right: Input, Baseline, Baseline + TSRA, Baseline + TSSRA, Baseline + GA, and Baseline + GA + RoIG.}
\label{fig4}
\end{figure*}

\begin{figure}[t]
\centering
\includegraphics[width=0.8\linewidth]{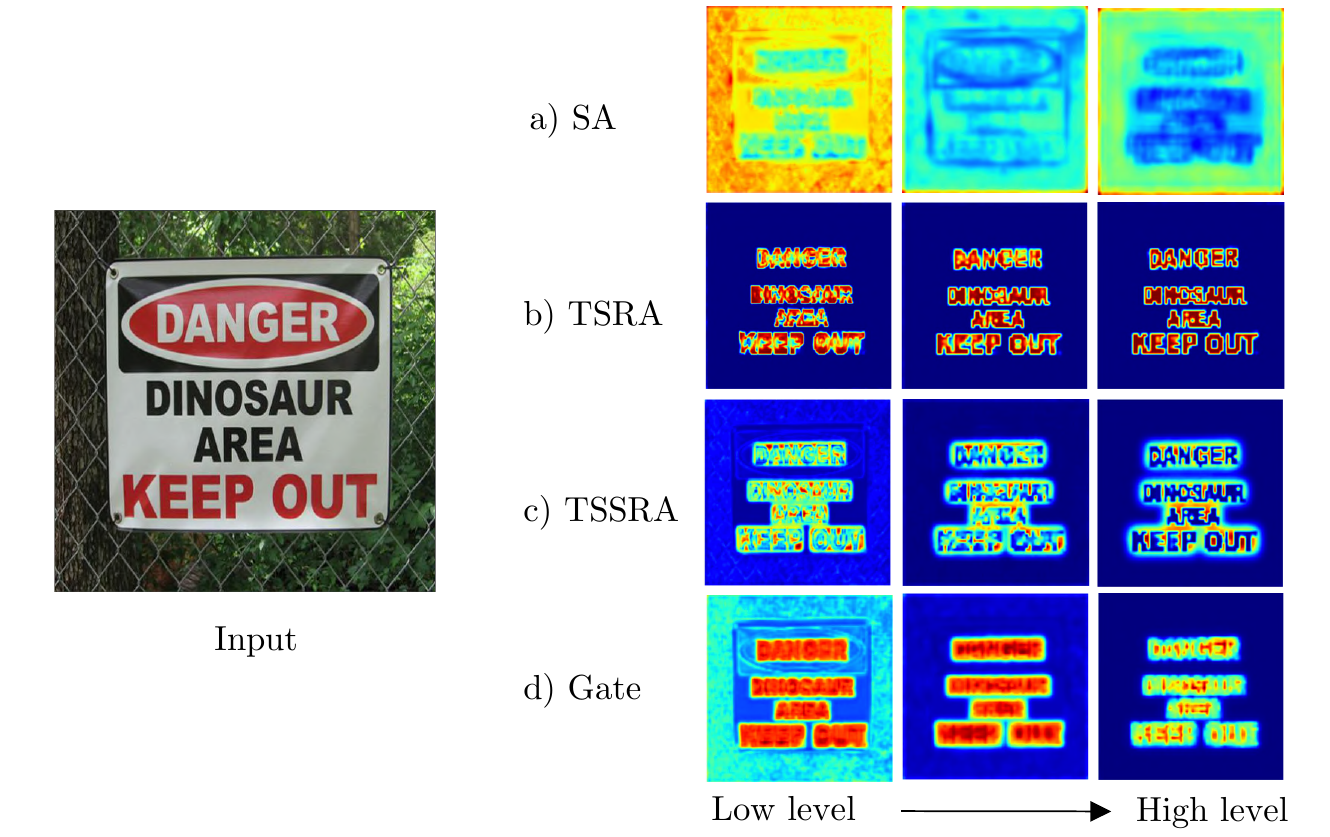} 
\caption{Visualization of the attention masks for differing encoding layers. The images progressively go from low-level features to high-level features. (a) is a visualization of simple spatial attention \cite{woo2018cbam}, (b) and (c) is a visualization of the text stroke and the surrounding attention that the GA generated. (d) is a visualization of how the GA used the gate parameter to aggregate TSA and TSSA. Spatial attention does a poor job finding the text strokes and the surrounding regions if simply applied in STR. In comparison, we can see that the GA pays more attention to the text stroke regions in low-level features while paying more attention to the surrounding regions of the text in high-level features.}
\label{fig5}
\end{figure}

\textbf{Attention}.
First, we performed the following three experiments to observe the effect of each attention on the STR results: Simple Attention (SA) \cite{woo2018cbam}, Text Stroke Region Attention (TSRA), and Text Stroke Surrounding Region Attention (TSSRA).

\Cref{table5} shows how the application of only SA does not improve the quality of STR.
\Cref{fig5}’s (a) shows how the application of only SA in STR does not lead to localization of the text stroke and the surrounding regions of the text stroke properly. However, both the TSRA and the TSSRA obtained higher PSNR and SSIM results. 
In particular, the evaluation result of the TSSRA was significantly better than the TSRA in Detection Eval. This shows that TSSR is more important. \Cref{fig4} demonstrates that results produced by the TSSRA reduce more artifacts than the TSRA. TSRA helps locate the TSR because it is a target for inpainting, but is inappropriate when focusing on the TSSR to erase text. On the other hand, TSSRA is appropriate to focus on the TSSR, and the model utilizes this to fill the TSR and generate higher quality output.

Furthermore, We found that having the GA module picking optimal ratios from an ensemble of both TSRA and TSSRA, then aggregating the features as appropriate was super effective. As \Cref{table5} and \Cref{fig4} clearly demonstrate, the evaluation scores of STR were highest when using the GA while also leaving almost no artifacts behind on the resulting image. In \Cref{fig5}, the GA module puts more emphasis on the surrounding regions of text strokes rather than the text strokes as it approaches higher-level features.

\textbf{Region of Interest Generation}.
In order to measure the effects of RoIG, we performed experiments with and without its use. \Cref{table5} demonstrates that RoIG significantly improves the quality of STR in all metrics. \Cref{fig4} shows that models with the application of RoIG left almost no artifacts with the best overall results. 

\begin{figure*}[t]
\centering
\includegraphics[width=0.9\textwidth]{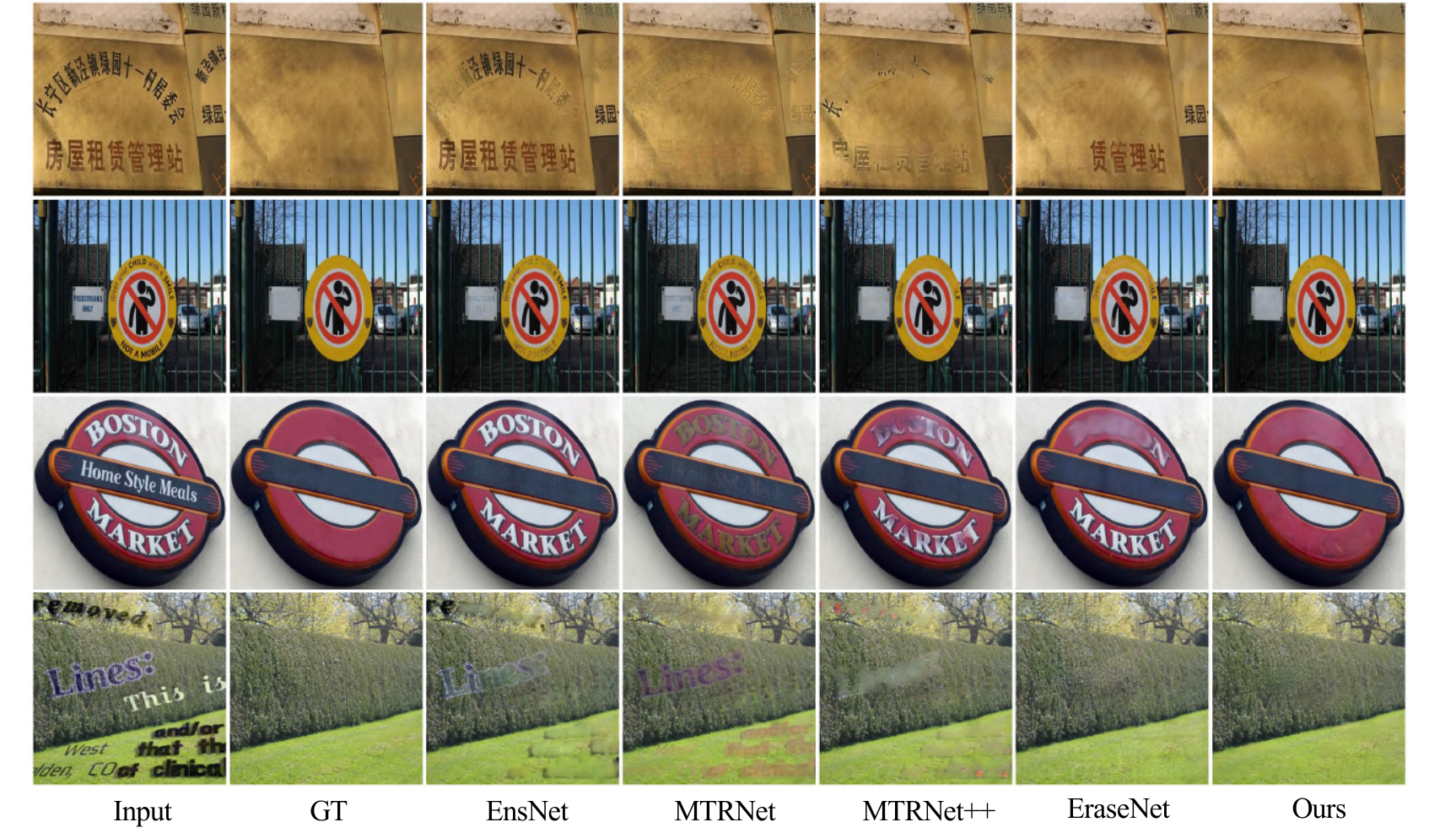} 
\caption{Comparison of output image results with other prominent STR models. Image from left to right: Input, Ground Truth, EnsNet \cite{zhang2019ensnet}, MTRNet \cite{tursun2019mtrnet}, MTRNet++ \cite{tursun2020mtrnet++}, EraseNet \cite{liu2020erasenet}, and Ours.}
\label{fig6}
\end{figure*}

\subsection{Comparison with previous methods}
As we mentioned in \Cref{modelanalysis}, the proposed model maintains a competitive edge with speed and model size while significantly improving the result quality than previous STR models on both real and synthetic data. The first row of \Cref{fig6} shows that EnsNet\cite{zhang2019ensnet} and EraseNet\cite{liu2020erasenet}, both of which do not use an explicit text box region, only partially erase text. The fourth row of the Figure shows that MTRNet\cite{tursun2019mtrnet} and MTRNet++\cite{tursun2020mtrnet++}  do not successfully remove all text from complex backgrounds without leaving behind artifacts or partially-erased text. However, our proposed model with GA and RoIG successfully outputs high-quality STR images without residual artifacts from images with small text, curved text, and text on complex backgrounds, even without additional refinement. 

\section{Conclusion}
Although there was a lot of progress in the STR area, it was difficult to establish the superiority of a model from the previously proposed methods because there was no standardized and fair way to evaluate performance. In this paper, we re-implemented prominent previously proposed methods, trained and evaluated on respective standardized datasets, and evaluated their accuracies, model size, and inference time in an objectively fair manner. 
We proposed a simple yet highly impactful STR method with Gated Attention (GA) and Region-of-Interest Generation (RoIG). GA uses attention on the text strokes and the surrounding region’s colors and textures to surgically erase text from images. RoIG makes the generator focus on only the region with text instead of the entire image for more efficient training. Our method significantly outperforms all existing state-of-the-art methods on all benchmark datasets in terms of inference time and output image quality.

\subsubsection{Acknowledgements.} We wish to thank Osman Tursun for providing codes of MTRNet and MTRNet++.

\clearpage
%
%
\bibliographystyle{splncs04}
\bibliography{egbib}

\newpage

\appendix

\section{Appendix}

\begin{figure}[t]
\centering
\includegraphics[width=1\linewidth]{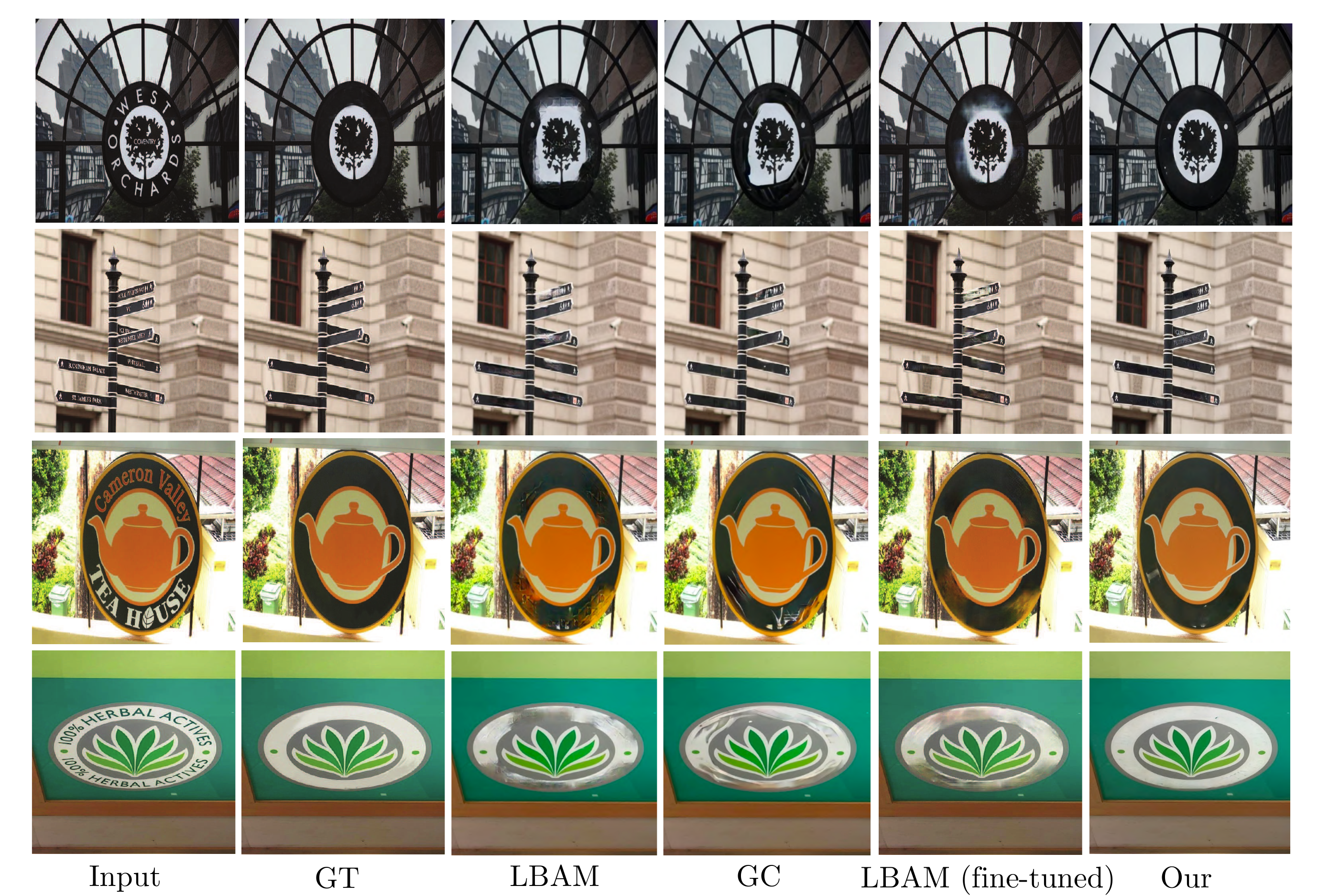} 
\caption{Comparison of the quality of images. Image from left to right: input image, ground truth image, LBAM pre-trained on Paris Street View, GC pre-trained on Places2, LBAM fine-tuned on our dataset, Ours.}
\label{fig7}
\end{figure}

\subsection{Comparison with general image inpainting}

We compared our proposed method with general inpainting methods. LBAM \cite{xie2019image}, Gated Convolution \cite{yu2019free} are adopted for comparison. We trained LBAM \cite{xie2019image}, which is pre-trained on Paris Street View \cite{doersch2015makes}, for 4 epochs on our combined dataset. For general inpainting methods, the text-stroke which is located outside of the box mask can affect the model's performance. To solve this issue, we provided the bounding box information, which was dilated 4 times with a 3x3 kernel, to the inpainting model. However, due to time constraints, we could not train GC \cite{yu2019free} on our combined dataset. Instead, we evaluated model, pre-trained on Places2 \cite{zhou2017places}, which has Contextual Attention module \cite{song2018contextual}. The quantitative results are shown in Tab. 6. For comparison, we used composited images generated by using box masks. \Cref{table6} shows that our proposed method outperforms the existing inpainting method in all metrics. The qualitative results are shown in \Cref{fig7}. As shown in \Cref{fig7}, the results of LBAM \cite{xie2019image} are blurry and incomplete. When masked regions contain complex backgrounds, the model can not reconstruct non-text information in masked regions properly, while our methods can reconstruct non-text regions and only erase text-stroke regions.

We acknowledge the lack of comparison with GC \cite{yu2019free}. We planned to trained GC \cite{yu2019free} on our combined dataset in the future. In addition, we planned to apply the Gated Convolution module to our model to compare performance between Gated Convolution and our proposed module without the influence of Contextual Attention \cite{song2018contextual}.

\begin{table}
\centering
\caption{Comparison for SCUT-EnsText (Image Eval). For a fair comparison, we provided box masks, which were dilated four times with a 3x3 kernel, to our methods. }
{\footnotesize
\begin{tabular}{l|l|c|ccc}
\hline
\multirow{2}{*}{Method} & \multirow{2}{*}{data} & Input & \multicolumn{3}{c}{SCUT-EnsText}  \\
     &  & size  & PSNR & SSIM & AGE  \\
\hline
LBAM \cite{xie2019image} & pre-trained & 512 & 34.20 & 96.13  & 1.6670 \\
 & pre-trained + Ours & 512 & 36.76 & 97.55  & 1.1404 \\
GC \cite{yu2019free} & pre-trained & 512 & 34.24 & 96.46  & 1.5049 \\
\hline
Ours & Ours & 512  & 39.20 & 98.11 & 0.8302 \\
\hline
\end{tabular}
}
\label{table6}
\end{table}
\setlength{\tabcolsep}{1.4pt}

\appendix

\end{document}